\documentclass[letterpaper, 10 pt, conference]{ieeeconf}  

\IEEEoverridecommandlockouts
\usepackage{graphics} 
\usepackage{epsfig} 
\usepackage{mathptmx} 
\usepackage{times} 
\usepackage{amsmath} 
\usepackage{amssymb}  
\usepackage{color}
\graphicspath{ {figures/} }
\usepackage[space]{cite}

 \usepackage[hidelinks]{hyperref}
\usepackage[algoruled]{algorithm2e}
\SetAlFnt{\small}
\usepackage{multirow}
\usepackage{multicol}
\usepackage{balance}

\title{\LARGE \bf
Tuning-Free Contact-Implicit Trajectory Optimization
}

\author{Aykut \"{O}zg\"{u}n \"{O}nol$^{1}$, Radu Corcodel$^{2}$, Philip Long$^{3}$, and Ta\c{s}k\i n Pad\i r$^{1}$
\thanks{This material is partially based upon work supported by National Science Foundation under Grant Nos. 1451427, 1544895, 1928654. The contribution outlined in this paper was implemented while A.{\"{O}}. {\"{O}}nol was an intern at Mitsubishi Electric Research Labs, and R. Corcodel's work was fully supported by Mitsubishi Electric Research Labs. The authors would like to thank Scott Jordan of UMass Amherst and Carlos Jose Nohra Khouri of Carnegie Mellon University for helpful discussions.}
\thanks{$^{1}$Institute for Experiential Robotics, Northeastern University, Boston, MA $^{2}$Mitsubishi Electric Research Labs, Cambridge, MA $^{3}$Irish Manufacturing Research, Dublin, Ireland
        {\tt\small \{onol.a, t.padir\}@northeastern.edu, corcodel@merl.com, philip.long@imr.ie}}%
}

\begin{document}

\maketitle
\thispagestyle{empty}
\pagestyle{empty}

\begin{abstract}
We present a contact-implicit trajectory optimization framework that can plan contact-interaction trajectories for different robot architectures and tasks using a trivial initial guess and without requiring any parameter tuning. This is achieved by using a relaxed contact model along with an automatic penalty adjustment loop for suppressing the relaxation. Moreover, the structure of the problem enables us to exploit the contact information implied by the use of relaxation in the previous iteration, such that the solution is explicitly improved with little computational overhead. We test the proposed approach in simulation experiments for non-prehensile manipulation using a 7-DOF arm and a mobile robot and for planar locomotion using a humanoid-like robot in zero gravity. The results demonstrate that our method provides an out-of-the-box solution with good performance for a wide range of applications.
\end{abstract}

\section{Introduction}
The introduction of contacts into a trajectory optimization problem leads to non-smooth dynamics and thus precludes the use of gradient-based optimization methods in a variety of robot manipulation and locomotion tasks. Therefore, much work has focused on mitigating the discrete nature of contacts by developing appropriate models which enable the optimization to reason about contacts. For this purpose, \cite{yunt2005trajectory,posa2014direct} propose to solve a nonlinear program with complementarity constraints. In \cite{gabiccini2018computational}, complementarity- and penalty-based contact models are used in a similar numerical scheme. \cite{mastalli2016hierarchical,marcucci2017two} present hierarchical strategies to increase the computational efficiency through warm starting. In \cite{manchester2017variational,patel2019contact}, the complementarity condition is relaxed, and methods are presented to improve the integration accuracy of the dynamics. Mordatch et al. \cite{todorov2012discovery,todorov2012manipulation} solve a convex program with soft constraints that model the dynamics with contacts by compromising physical realism. In \cite{todorov2012synthesis}, a time-stepping scheme with a smoother variant of the complementarity constraints is proposed, and iterative linear quadratic regulator (iLQR) \cite{todorov2004ilqr} is employed to solve the problem near real-time. Similarly, \cite{hutter2018trajectory} proposes a bi-level optimization based on iLQR with implicit hard-contact constraints. iLQR is also used in \cite{buchli2017discovery,buchli2018npmc} but with an explicit smooth contact model, in which the contact force is a function of the distance, so that dynamic motions for a quadruped robot can be planned in real-time.

\begin{figure}
  \centering
  \includegraphics[width=0.72\columnwidth]{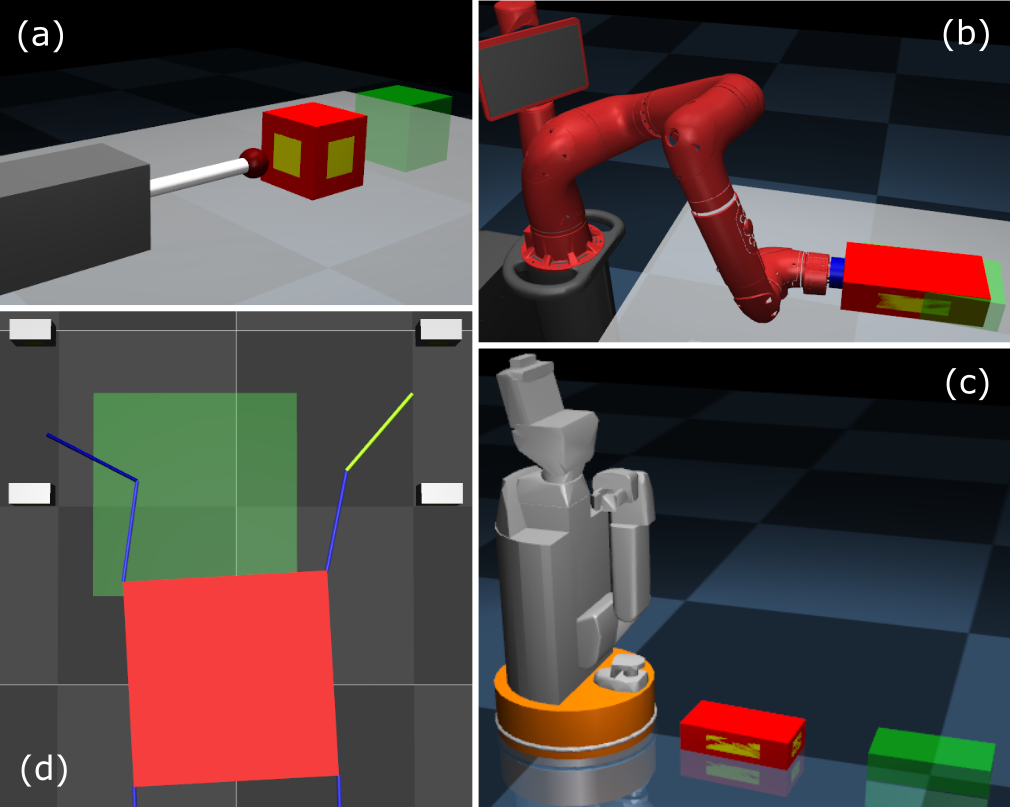}
  \caption{Applications considered in this study: (a) 1-DOF pusher-slider system for a visual analysis of the problem, (b) a 7-DOF robot arm pushing a box, (c) a mobile robot pushing a box, and (d) locomotion in zero gravity.}
  \label{fig:applications}
\vspace{-0.12in}
\end{figure}

While smooth models facilitate convergence, they also lead to  physical inaccuracies and are quite difficult to tune.  With this in mind, we previously proposed a variable smooth contact model (VSCM) \cite{onol2018comparative} in which virtual forces acting at a distance are injected to the underactuated dynamics with rigid-body contact mechanics. The  virtual  forces  are exploited to discover contacts and  minimized  throughout the optimization. Consequently, physically-feasible motions are obtained while maintaining fast convergence. Additionally, shooting methods, such as iLQR, are typically sensitive to the initial guess \cite{betts1998survey}; thus, in \cite{onol2019contact}, we proposed a variant of the successive convexification algorithm that was originally proposed in \cite{acikmese2016scvx,acikmese2018scvx} and showed that it outperforms iLQR for a contact-implicit trajectory optimization (CITO) problem. The use of the VSCM and the successive convexification algorithm (SCVX) significantly mitigates the sensitivity to the initial guess and the burden of tuning by reducing the number of tuning parameters to one, namely a penalty on the virtual stiffness. Nevertheless, it may be required to tune this penalty when the task or the robot is changed; and without extra tuning, abrupt changes may occur in the planned motions even with minor task modifications. Moreover, the resulting contacts are usually impulsive due to the structure of the contact model.

In order to address these issues, we introduce a penalty loop approach for CITO that is analogous to state-of-the-art trajectory optimization methods for collision avoidance such as TrajOpt \cite{schulman2014motion} and GuSTO \cite{bonalli2019gusto}. In these methods, the penalty on constraints is gradually increased so that the optimization can be initialized with an infeasible trajectory that is in collision. The robot links are eventually pulled out of collisions by following the gradients. For the CITO problem, an infeasible trajectory corresponds to a motion that completes the task by using non-physical forces that act from a distance, namely the virtual forces. Hence, we develop a method that automatically adjusts the penalty on the relaxation parameters until a motion that completes the task using only physical forces is found. In the CITO case, the solution can be further improved by explicitly exploiting the contact information from the relaxation. In other words, the residual virtual forces indicate the position, time, and magnitude of contact forces required to complete the task. For this purpose, we develop a computationally-cheap post-processing step that improves the solution.

We consider non-prehensile manipulation applications using a 1-degree-of-freedom (DOF) pusher, a 7-DOF arm, and a holonomic mobile robot and a planar locomotion application in zero gravity, see Fig.~\ref{fig:applications}. We test the proposed approach for various goal positions to demonstrate the robustness of our framework. In all cases, the exact same configuration of the pipeline is used with a trivial initial guess, in which the robot stands still. To the best of our knowledge, this is the first attempt to generalize CITO which is the main contribution of this paper to the theory of optimization-based planning of contact-interaction trajectories.


\section{Methodology}
\subsection{Dynamic Model}
The dynamics of an underactuated system with $n_a$ actuated DOF and $n_u$ unactuated DOF, subject to external forces due to frictional rigid-body contacts and virtual forces generated by the contact model is given by
\begin{equation}
     \mathbf{M}(\mathbf{q}) \ddot{\mathbf{q}} + \mathbf{c}(\mathbf{q},\dot{\mathbf{q}}) = \mathbf{S}_a^T \boldsymbol{\tau} + \mathbf{J}^T_{c}(\mathbf{q}) \boldsymbol{\lambda}_{c} + 
     \mathbf{S}_u^T \boldsymbol{\lambda}_{v},
     \label{eq:dynamics}
\end{equation}
where $\mathbf{q} \triangleq [\mathbf{q}_a^T, \mathbf{q}_u^T]^T\in\mathbb{R}^{n_a+n_u}$ is the configuration vector; $\mathbf{M}(\mathbf{q}) \in \mathbb{R}^{(n_a+n_u) \times (n_a+n_u)}$ is the mass matrix; $\mathbf{c}(\mathbf{q},\dot{\mathbf{q}}) \in \mathbb{R}^{n_a+n_u}$ represents the Coriolis, centrifugal, and gravitational terms; $\mathbf{S}_a=[\mathbb{I}_{n_a \times n_a} \ \mathbf{0}_{n_a \times n_u}]$ is the selection matrix for the actuated DOF and $\mathbf{S}_u=[\mathbf{0}_{n_u \times n_a} \ \mathbb{I}_{n_u \times n_u}]$ is the selection matrix for the unactuated DOF; $\boldsymbol{\tau} \in \mathbb{R}^{n_a}$ is the vector of generalized joint forces; $\boldsymbol{\lambda}_{c} \in \mathbb{R}^{6n_{c}}$ is the vector of generalized contact forces at $n_c$ contact points and $\mathbf{J}_{c}(\mathbf{q}) \in \mathbb{R}^{6n_{c} \times (n_a+n_u)}$ is the Jacobian matrix mapping the joint velocities to the Cartesian velocities at the contact points and $\boldsymbol{\lambda_{v}}\in \mathbb{R}^{n_u}$ is the vector of generalized contact forces on the unactuated DOF generated by the contact model. In this paper, for $n_f$ free bodies in $SE(3)$ (\textit{e.g.} objects or the torso of a humanoid), we set $n_u=6n_f$. The state of the system is represented by $\mathbf{x} \triangleq [\mathbf{q}^T \ \mathbf{\dot{q}}^T]^T \in \mathbb{R}^{n}$ where $n=2(n_a+n_u)$.

It should be noted that there are two types of contact mechanics in this system: (\textit{i}) frictional contact forces due to physical contacts in the simulated world (\textit{i.e.} contacts detected by the physics engine) which are effective on all DOF; and (\textit{ii}) virtual forces due to the contact model which are exerted only on the unactuated DOF, as explained in \ref{subsec:contact}.

In this study, the generalized joint forces are decomposed as $\boldsymbol{\tau}=\boldsymbol{\tau}_u+\Tilde{\mathbf{c}}-\Tilde{\mathbf{J}}^T_{c}\Tilde{\boldsymbol{\lambda}}_{c}$, where $\Tilde{\mathbf{c}}$, $\Tilde{\mathbf{J}}_{c}$, and $\Tilde{\boldsymbol{\lambda}}_{c}$ are the estimations of $\mathbf{c}(\mathbf{q},\mathbf{\dot{q}})$, $\mathbf{J}_{c}(\mathbf{q})$, and $\boldsymbol{\lambda}_{c}$; and $\boldsymbol{\tau}_u\in\mathbb{R}^{n_a}$ is the vector of control variables associated with the joint forces. This helps to center the optimization problem in terms of the joint forces by directly relating the control term to the acceleration.

\subsection{Contact Model}\label{subsec:contact}
The contact model considers $n_{p}$ pairs of predefined contact candidates on the robot (\textit{e.g.} end-effector links) and in the environment (\textit{e.g.} the surfaces of an object). For each contact pair, the magnitude of the virtual force normal to the surface $\gamma\in \mathbb{R}_+$ is calculated by $\gamma(\mathbf{q}) = k e^{-\alpha \phi(\mathbf{q})}$ using the virtual stiffness $k \in \mathbb{R}_+$, the signed distance between the contact candidates $\phi\in \mathbb{R}$, and the curvature $\alpha \in \mathbb{R}_+$. The corresponding generalized virtual force acting on the free body associated with the contact pair $\boldsymbol{\lambda}_v \in \mathbb{R}^6$ is calculated by $\boldsymbol{\lambda}_v(\mathbf{q}) = \gamma(\mathbf{q})[\mathbb{I}_3 \ -\mathbf{\hat{l}}(\mathbf{q})]^T\mathbf{n}(\mathbf{q})$, where $\mathbb{I}_3$ is $3\times3$ identity matrix, $\mathbf{l}$ is the vector from the center of mass of the free body to the nearest point on the contact candidate on the robot, $\hat{\mathbf{l}}$ is the skew-symmetric matrix form of $\mathbf{l}$ that performs the cross product, and $\mathbf{n} \in \mathbb{R}^3$ is the contact surface normal. The net virtual force acting on a free body is the sum of the virtual forces corresponding to the contact candidates associated with that body. As a result, the virtual forces provide a smooth relationship between the dynamics of the free bodies and the configuration of the system.

In the VSCM, the virtual stiffness values $\mathbf{k}\in\mathbb{R}^{n_p}$ are decision variables of the optimization. Thus, the vector of control variables is $\mathbf{u} \triangleq [\boldsymbol{\tau}_u^T, \ \mathbf{k}^T]^T \in \mathbb{R}^m$ where $m=n_a+n_p$.

\subsection{Trajectory Optimization Problem}
A finite-dimensional trajectory optimization problem for $N$ time steps can be written in terms of state and control trajectories $\mathbf{X} \triangleq [\mathbf{x}_1,...,\mathbf{x}_{N+1}]$ and $\mathbf{U} \triangleq [\mathbf{u}_1,...,\mathbf{u}_{N}]$; final and integrated cost terms $C_F$ and $C_I$; and lower and upper control and state bounds $\mathbf{u}_L$, $\mathbf{u}_U$, $\mathbf{x}_L$, and $\mathbf{x}_U$:
\begin{subequations}
    \begin{align}
        \underset{\mathbf{U}}{\text{minimize }} C(\mathbf{X},\mathbf{U}) \triangleq C_F(\mathbf{x}_{N+1}) + \sum_{i=1}^{N} C_I(\mathbf{x}_i,\mathbf{u}_i)
    \end{align}
subject to:
    \begin{align}
        & \mathbf{x}_{i+1} = f(\mathbf{x}_i,\mathbf{u}_i) \text{ for } i=1,...,N, \label{eq:dynamics_constraint} \\
        & \mathbf{u}_{L} \leq \mathbf{u}_{1,...,N} \leq \mathbf{u}_{U}, \ \mathbf{x}_{L} \leq \mathbf{x}_{1,...,N+1} \leq \mathbf{x}_{U},
    \end{align}
\end{subequations}
where $\mathbf{x}_{i+1} = f(\mathbf{x}_i,\mathbf{u}_i)$ describes the evolution of the nonlinear dynamics over a control period $i$.

Locomotion and non-prehensile manipulation tasks can be defined based on the desired torso/object configurations. In this study, we use a weighted quadratic final cost in terms of the deviations of the position and orientation of the free body from the desired pose, $p_e$ and $\theta_e$:
\begin{equation}
    C_F = w_1 p_e^2 + w_2 \theta_e^2, \label{eq:cf}
\end{equation}
where $w_1$ and $w_2$ are the weights. To suppress all virtual forces, the $L^1$-norm of the virtual stiffness variables (\textit{i.e.} an exact penalty method \cite{nocedal2006penalty}) is penalized in the integrated cost:
\begin{equation}
    C_I = \omega ||\mathbf{k}_i||_1. \label{eq:ci}
\end{equation}
The penalty $\omega$ is adjusted by the penalty loop algorithm, as described in \ref{subsec:PL}.

\subsection{Successive Convexification}
The original successive convexification algorithm proposed in \cite{acikmese2016scvx,acikmese2018scvx} is based on repeating three main steps in successions: (\textit{i}) linearizing non-convex constraints (\textit{e.g.} the nonlinear dynamics) about the trajectory from the previous succession, (\textit{ii}) solving the resulting convex subproblem in a trust region to avoid artificial unboundedness due to linearization, and (\textit{iii}) adjusting the trust-region radius based on the fidelity of the linear approximations.

The convex subproblem is given by:
\begin{subequations}
    \begin{align}
        &\underset{\mathbf{\delta x}_1,...,\mathbf{\delta x}_{N+1},\mathbf{\delta u}_1,...,\mathbf{\delta u}_{N}}{\text{minimize }} L \triangleq C(\mathbf{X}^s + \mathbf{\delta X},\mathbf{U}^s + \mathbf{\delta U})
    \end{align}
subject to:
    \begin{align}
        & \mathbf{\delta x}_{i+1} = \mathbf{A}_i \mathbf{\delta x}_i + \mathbf{B}_i \mathbf{\delta u}_i \text{ for } i=1,...,N,  \label{eq:linearized_dynamics_constraint}  \\
        & \mathbf{x}_{L} \leq \mathbf{x}_i^s + \mathbf{\delta x}_i \leq \mathbf{x}_{U} \text{ for } i=1,...,N+1, \\
        & \mathbf{u}_{L} \leq \mathbf{u}_i^s + \mathbf{\delta u}_i \leq \mathbf{u}_{U} \text{ for } i=1,...,N, \\
        & ||\mathbf{\delta X}||_1 + ||\mathbf{\delta U}||_1 \leq r^s,
    \end{align}
    \label{eq:qp}
\end{subequations}
where $(\mathbf{X}^s,\mathbf{U}^s)$ is the trajectory from the succession $s$; $\mathbf{A}_i \triangleq \partial f(\mathbf{x}_i,\mathbf{u}_i) / \partial \mathbf{x}_i |_{\mathbf{x}_i^s,\mathbf{u}_i^s}$; $\mathbf{B}_i \triangleq \partial f(\mathbf{x}_i,\mathbf{u}_i) / \partial \mathbf{u}_i|_{\mathbf{x}_i^s,\mathbf{u}_i^s}$; $\delta \mathbf{x}_i \triangleq \mathbf{x}_i - \mathbf{x}_i^s$; $\delta \mathbf{u}_i \triangleq \mathbf{u}_i - \mathbf{u}_i^s$; and $r$ is the trust-region radius. Additionally, virtual controls can be added to this problem to prevent artificial infeasibility due to linearization \cite{acikmese2018scvx}.


The convex subproblem is a simultaneous problem, and therefore has a larger size but a sparse structure, which can be exploited by a suitable solver. After solving the convex subproblem, we apply only the change of controls instead of applying the changes for both states and controls. The state trajectory is then recalculated by rolling-out the dynamics. This modification prevents the accumulation of defects (\textit{i.e.} $f(x_i,u_i) - x_{i+1}$) that may occur when using the penalty approach in the original method as well as improves the convergence speed by allowing larger trust regions in our experiments. As a result, the modified method combines the numerical efficiency of direct methods and the accuracy of shooting methods. Although it does not have a convergence proof, unlike the original algorithm, it is shown to provide fast and reliable convergence for CITO in \cite{onol2019contact}.

\subsection{Penalty Loop} \label{subsec:PL}

The SCVX is initialized with large virtual stiffness values to enable the optimization algorithm to reason about contacts. With judicious tuning, the virtual forces vanish yielding a motion that solves the task by using only physical contacts as the optimization converges. However, the penalty on the virtual stiffness $\omega$, plays an important role in this process. While a small penalty may result in physically-inconsistent motions due to leftover virtual forces, a motion that completes the task may not be found if the penalty is too large. Although the tuning of this penalty is fairly straightforward, it hinders the generalization of the method for a wide range of tasks and robots. In order to address this issue, we propose a penalty loop approach that adjusts the penalty similar to a subgradient method \cite{shor2012minimization}. Furthermore, the solution is improved after each iteration through a computationally-cheap post-processing stage by exploiting the contact information embedded in the stiffness variables.

The proposed penalty loop approach is summarized in Algorithm \ref{alg:penalty}. In the first iteration, the SCVX is run with a light penalty value. If the solution satisfies the pose constraints, the weight is increased to reduce the stiffness values; otherwise, the solution is rejected and the weight is reduced by the half of the previous change. Then, a post process is performed on the optimal trajectory to attract the robot links associated with the non-zero stiffness values towards the corresponding contact candidates in the environment using a pulling controller, which is outlined in \ref{subsec:post}. This process is carried out only if the average stiffness value is below a threshold, which usually holds after the first iteration in our experiments. The position constraint is normalized with respect to initial position error such that the method scales better to different orders of magnitude (meters for locomotion versus centimeters in the manipulation task). 

\begin{algorithm}[!bt]
    \SetAlgoLined
    \SetKwInOut{Input}{Input}
\SetKwInOut{Output}{Output}
    \SetKwInOut{Data}{Data}
    \Input{Initial state vector $\mathbf{x}_1$ and initial control trajectory $\mathbf{U}$.}
    \Output{Optimal state and control trajectories.}
    \Data{Initial penalty value $\omega^1>0$, penalty step size $\Delta \omega_s>0$, initial position error $p_e^0$, position tolerance $\epsilon_p>0$, rotation tolerance $\epsilon_{\theta}>0$, avg. stiffness threshold for post processing $k_{threshold}$, max. stiffness tolerance $\epsilon_k>0$.}
    $j=1$, $\mathbf{U}^j=\mathbf{U}$ \\
    \Repeat{$j>j_{max} \lor (k_{max} \leq \epsilon_k \land p_e/p_e^0 \leq \epsilon_p$)}
    {
        \textbf{Step 1} $(\mathbf{X},\mathbf{U}) \leftarrow$ SCVX$(\mathbf{U}^j,\omega^j$). \\
        \textbf{Step 2} Calculate position error $p_e$, rotation error $\theta_e$, max. and avg. stiffness values $k_{max}$ and $k_{avg}$. \\
        \eIf{$p_e/p_e^0 \leq \epsilon_p \land \theta_e \leq \epsilon_{\theta}$}
        {
            $\Delta \omega^j \leftarrow \Delta \omega_s$ \\
            $\omega^{j+1} \leftarrow \omega^j + \Delta \omega^j$ \\
            \If{$k_{avg}^j \geq k_{avg}^{j-1}$}
            {
                Reject the solution $\mathbf{U}^{j+1} \leftarrow \mathbf{U}^j$, $k_{avg}^j \leftarrow k_{avg}^{j-1}$, $j \leftarrow j+1$, and go back to \textbf{Step 1}.
            }
        }
        {
            $\Delta \omega^j \leftarrow - \Delta \omega^{j-1}/2$ \\
            $\omega^{j+1} \leftarrow \omega^j + \Delta \omega^j$ \\
            Reject the solution $\mathbf{U}^{j+1} \leftarrow \mathbf{U}^j$, $k_{avg}^j \leftarrow k_{avg}^{j-1}$, $j \leftarrow j+1$, and go back to \textbf{Step 1}.
        }
        \eIf{$k_{avg}<k_{threshold}$}
        {
                    \textbf{Step 3} Apply the pulling controller:\\
        $(\mathbf{X}_{pp},\mathbf{U}_{pp}) \leftarrow$ PC$(\mathbf{x}_1,\mathbf{U})$.\\
        \textbf{Step 4} Perform the hill-climbing search:\\ $(\mathbf{X}_{pp},\mathbf{U}_{pp}) \leftarrow$ HCS$(\mathbf{x}_1,\mathbf{U}_{pp}^j)$.\\
        \textbf{Step 5} Recalculate $p_e$.\\
        \eIf{$p_e/p_e^0\leq\epsilon_p$}
        {
            Accept the post-processing step: $(\mathbf{X}^{j+1},\mathbf{U}^{j+1}) \leftarrow (\mathbf{X}_{pp},\mathbf{U}_{pp})$.
        }
        {
            Reject the post-processing step: $(\mathbf{X}^{j+1},\mathbf{U}^{j+1}) \leftarrow (\mathbf{X},\mathbf{U})$.
        }
        }
        {Accept the solution: $(\mathbf{X}^{j+1},\mathbf{U}^{j+1}) \leftarrow (\mathbf{X},\mathbf{U})$
        }

        $j \leftarrow j+1$
     }
    \KwRet{$(\mathbf{X}^{j},\mathbf{U}^{j})$.}
    \caption{Penalty Loop}
    \label{alg:penalty}
\end{algorithm}

\subsection{Post Process} \label{subsec:post}
After solving the SCVX at each penalty loop iteration, the residual virtual stiffness variables indicate the position, timing, and magnitude of forces required to complete the task. To exploit this information, we develop a controller that pulls the contact candidates on the robot associated with non-zero stiffness variables towards the corresponding contact candidates in the environment.

For a contact pair $p$ and a control period $i$, the pulling force $\mathbf{f}[p,i]\in \mathbb{R}^3$ is calculated from the distance vector $\mathbf{d}[p,i]\in \mathbb{R}^3$ and the associated virtual stiffness value $k[p,i]$:
\begin{equation}
    \mathbf{f}[p,i] = k[p,i] \mathbf{d}[p,i].
\end{equation}
Here, $\mathbf{d}$ is the vector from the center of mass of the contact candidate on the robot to the point that is offset from the center of the contact candidate in the environment. The offset is initialized at $d_0$ for the first penalty iteration and reduced for the following iterations by dividing it by the number of successful penalty iterations. This offset helps reaching occluded surfaces in the environment. Alternatively, a potential field approach may be used with repulsive forces on the surfaces with zero stiffness values.

The corresponding generalized joint force vector $\tau_{pull}[i]\in \mathbb{R}^{n_a}$ is calculated by:
\begin{equation}
    \boldsymbol{\tau}_{pull}[i] = \textstyle{\sum_{p=1}^{n_p}{\mathbf{J}_t^T[p,i] \mathbf{f}[p,i]}},
\end{equation}
where $\mathbf{J}_t[p,i]\in \mathbb{R}^{3\times n_a}$ is the translational Jacobian matrix for the center of mass of the contact candidate on the robot.

To prevent the pulling force generating abrupt motions for large stiffness values, a damping controller is applied to keep the joint velocities close to the planned motion:
\begin{equation}
    \boldsymbol{\tau}_{damp}[i] = \mathbf{K}_v \mathbf{S}_a^T \mathbf{M}(\mathbf{q}[i]) \mathbf{\dot{q}}_e[i],
\end{equation}
where $\mathbf{K}_v\in\mathbb{R}^{n_a \times n_a}$ is a positive-definite gain matrix, $\mathbf{\dot{q}}_e$ is the deviation of the joint velocities from the planned velocities and $\boldsymbol{\tau}_{damp}[i] \in \mathbb{R}^{n_a}$ is the generalized joint forces for damping. This computation can be done very efficiently by using the sparse form of the inertia matrix.

In Algorithm \ref{alg:penalty}, applying these two steps is referred as the pulling controller (PC). The inputs of the PC are the initial state vector and the optimal control trajectory obtained from the SCVX. The PC adds the pulling and damping forces to the planned joint forces, \textit{i.e.} $\boldsymbol{\tau} = \boldsymbol{\tau} + \boldsymbol{\tau}_{pull} + \boldsymbol{\tau}_{damp}$, and outputs the resulting control and state trajectories. The PC attracts virtually active robot links to the corresponding contact candidates in the environment to facilitate physical contacts. However, as the distances get smaller, the planned stiffness values may lead to excessively large virtual forces. To prevent that, we perform a naive hill-climbing search (HCS) after applying the PC. In this step, non-zero stiffness values are reduced by the change of the final cost divided by the previous change as long as the nonlinear pose error decreases. This step also helps to suppress virtual forces explicitly, \textit{i.e.} independently of the penalty increase.

\section{Applications}
\subsection{1-DOF Pusher-Slider System}
For a visual analysis of the problem, we evaluate the proposed approach for a simplistic problem: a 1-DOF pusher-slider system with a single time step of 1 s, see Fig.~\ref{fig:applications}(a). The task is to push the slider 20~cm forward (Task 1a), and there is only one contact pair that consists of the tip of the pusher and the front face of the slider.

\subsection{Pushing with a 7-DOF Arm}
Similar to \cite{onol2018comparative,onol2019contact,sleiman2019contact}, we test the method for pushing a box on a table, see Fig.~\ref{fig:applications}(b). For this application, we consider a 7-DOF Sawyer Robot by Rethink Robotics. In addition to pushing the box forward, which is the only task considered in \cite{onol2018comparative,onol2019contact,sleiman2019contact}, the method is evaluated for side and diagonal pushes as well, which are typically more challenging as the robot must use the occluded faces of the object. In this environment, there are four contact pairs between the side faces of the box and the cylindrical end-effector flange of the robot. The simulation time is 1 s, and the control sampling period is 0.1 s. Three forward pushing tasks are considered to move the box 5 cm, 10 cm, and 30 cm (\textit{i.e.} Tasks 1b, 2b, and 3b). The goal of these tasks is to show that, even with identical initialization conditions, the method performs reliably for diverse problems, \textit{i.e.} tasks that require gentle contact interactions for slight motions or impulsive motions to move the object out of the robot's workspace. Moreover, we evaluate the method to push the box 10 cm left and right (Tasks 4b and 5b) as well as for a diagonal push to move the box 20 cm forward and 20 cm left (Task 6b).

\subsection{Pushing with a Mobile Robot}
Another application studied here is non-prehensile manipulation with a mobile robot. We consider a Human Support Robot (HSR) by Toyota pushing a box using its velocity-controlled, holonomic base, as shown in Fig.~\ref{fig:applications}(c). There are four contact pairs between the side faces of the box and the cylindrical base of the robot. As the translational and rotational velocities are bounded by $\pm$2 m/s and $\pm$2 rad/s, a simulation time of 5 s and a control sampling period of 0.5 s are used for this application. A forward pushing task to move the box 50 cm (Task 1c) and two diagonal pushing tasks are considered. It is observed that when the default friction coefficient of the physics engine ($\mu=1$) is used, the robot heavily relies on the frictional forces for the diagonal pushes, which seems unrealistic. In order to show that the method is capable of avoiding that as well, we repeat these tasks using $\mu=0.1$. Hence, Tasks 2c and 3c require to move the box 20 cm forward and 20 cm left, and Tasks 4c and 5c require to move the box 30 cm forward and 10 cm right.

\subsection{Planar Locomotion in Zero Gravity}
Lastly, the proposed framework is tested for a locomotion application to demonstrate that it can make and break multiple contacts simultaneously. We consider a planar, humanoid-like robot with a prismatic torso and 2-DOF cylindrical arms and legs. The environment has zero gravity which avoids stability constraints allowing exact utilization of the proposed framework. The task is specified in terms of the torso's desired pose  which the robot can reach by using four static bricks in the environment, as shown in Fig.~\ref{fig:applications}(d). However, as the motion is undamped without contacts, the robot must also use contacts to slow down or stop. In this case, there are 8 contact candidates in the environment that are the front and rear faces of the bricks, and 4 contact candidates on the robot that are the end links of the arms and the legs. These candidates are paired based on the sides, so that there are 16 contact candidates in total. Indeed, the legs are never used in the tasks considered here but are added in order to show that the proposed method can immediately reject superfluous virtual stiffness variables as well as its capability of handling many contact pairs. For this application, the simulation time is 2 s, and the control sampling period is 0.1 s. We consider two tasks to move the torso forward 70 cm and 100 cm, \textit{i.e.} Tasks 1d and 2d. Tasks 3d, 4d, 5d, and 6d require to move the torso 70 cm forward and 20 cm right, 100 cm forward and 20 cm left, 90 cm forward and 25 cm left, and 40 cm forward and 20 cm right, respectively.

\section{Simulation Experiments}
\subsection{Software Implementation}
We simulate the dynamics in MuJoCo \cite{mujoco} since it is found to be advantageous for rigid-body dynamics with contacts \cite{mujoco_comparison} and employs a smooth contact model that facilitates the gradient-based optimization \cite{mujoco_contact}. The large-scale, sparse solver SQOPT \cite{sqopt77} is used to solve the convex subproblem in \eqref{eq:qp} which has a large and sparse equality constraint \eqref{eq:linearized_dynamics_constraint} and a sparse quadratic cost with only 6 non-zero elements. The distance between contact candidates and the nearest point on the robot are calculated by the Gilbert-Johnson-Keerthi algorithm \cite{gjk} implemented in FCL \cite{fcl}. The derivatives of the dynamics are approximated by central differences.\footnote{The code is available at \href{https://www.merl.com/research/license\#CITO}{https://www.merl.com/research/license\#CITO}.} However, it is possible to make this process more computationally efficient and accurate using the analytical derivatives developed in \cite{carpentier2018analytical,pinocchio}. Computations are run on a workstation with Intel Core i7-6700K processor.

\subsection{Parameter Values}
In all cases, the same parameter values and initial seed are used. The weights in the final cost \eqref{eq:cf} are $w_1=10^4$ and $w_2=1$. We use a trivial initial guess with zero joint force values $\boldsymbol{\tau}_u=\mathbf{0}$ (\textit{i.e.} the robot does not move) and large stiffness values $k=10$~N/m. The upper bound for the stiffness variables is 20~N/m and $\alpha=10$. In the penalty loop algorithm, the initial penalty value $\omega_s=0.1$, the penalty step size $\Delta \omega_s=1.5$, and the average stiffness threshold is 2~N/m. The tolerance values are selected as $\epsilon_p=30\%$, $\epsilon_{\theta}=1$ rad, and $\epsilon_{k}=0.1$~N/m, yet they can be tightened as needed for the application. For the post-process stage, $\mathbf{K}_v=2.5\mathbb{I}$ and the initial offset distance, $d_0$ is arbitrarily selected as 5~cm but the procedure is not sensitive to this value. In the HCS, the initial step size is 0.1~N/m, and the step size and cost tolerances are $10^{-3}$ are $10^{-2}$.

\subsection{Visual Analysis}
\begin{figure}
  \centering
  \includegraphics[width=0.7\columnwidth]{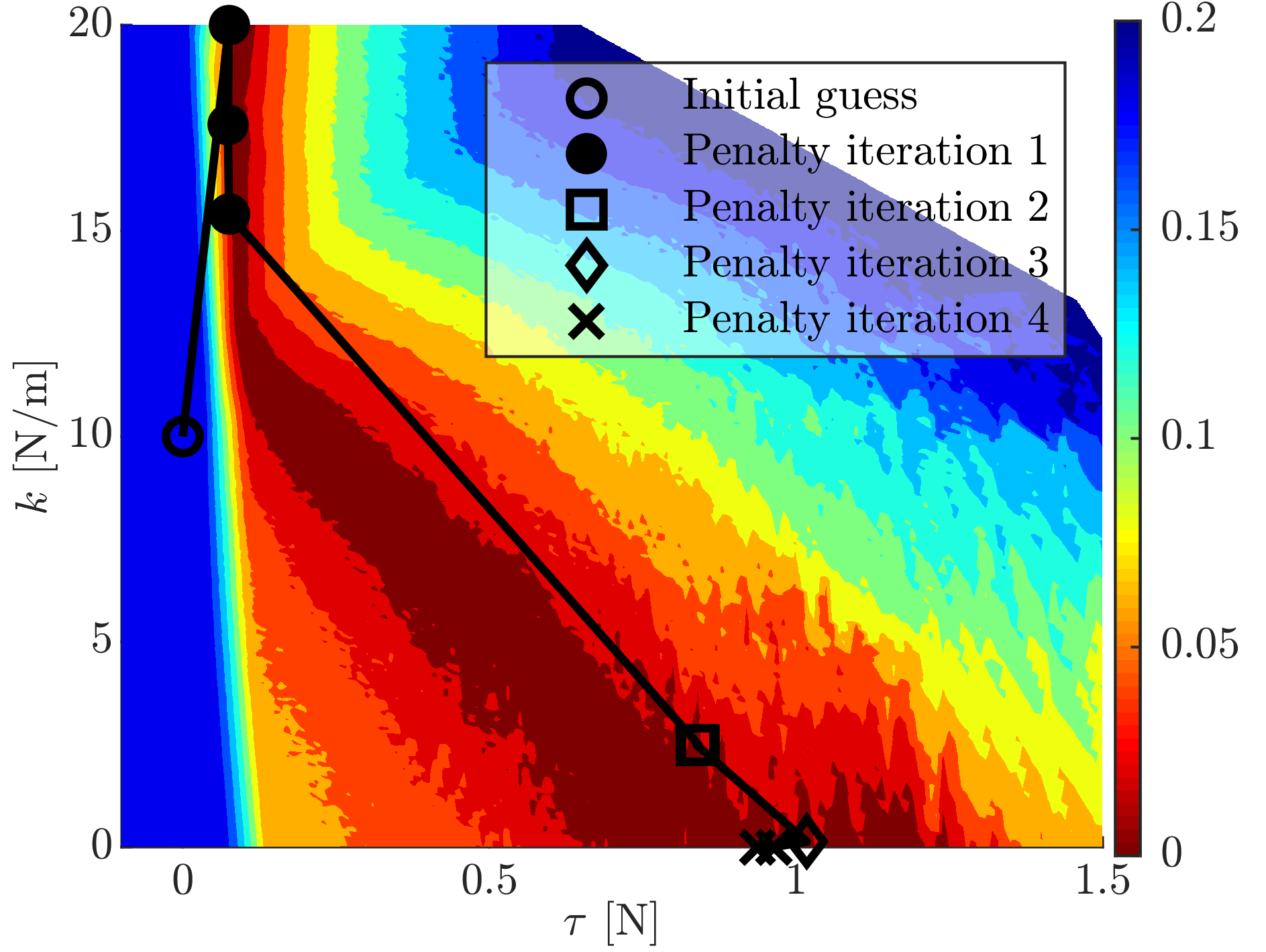}
  \caption{Change of the cost function over control variables for the 1-DOF pusher-slider system and the progress of the penalty loop algorithm.}
  \label{fig:contour_progress}
  \vspace{-0.12in}
\end{figure}

Figure \ref{fig:contour_progress} depicts the cost function with respect to the control variables as well as the progress of the proposed method for the 1-DOF pusher-slider system. The large blue band shows that the cost is mostly flat when the physical and virtual forces are zero. In other words, it is difficult for the gradient-based optimization to find a motion that completes the task without a reasonable initial guess. However, the virtual force relaxes the problem by providing steep gradients from the trivial initial guess, and the SCVX accomplishes the task using mostly the virtual force in the first penalty iteration. Moreover, after each iteration the solution is improved substantially by the pulling controller that gets the robot closer to the box and by reducing the virtual stiffness using the HCS. Consequently, a motion that satisfies the task constraints by using only physical forces is obtained.

\subsection{Results}
\begin{figure}
  \centering
  \includegraphics[width=.85\columnwidth]{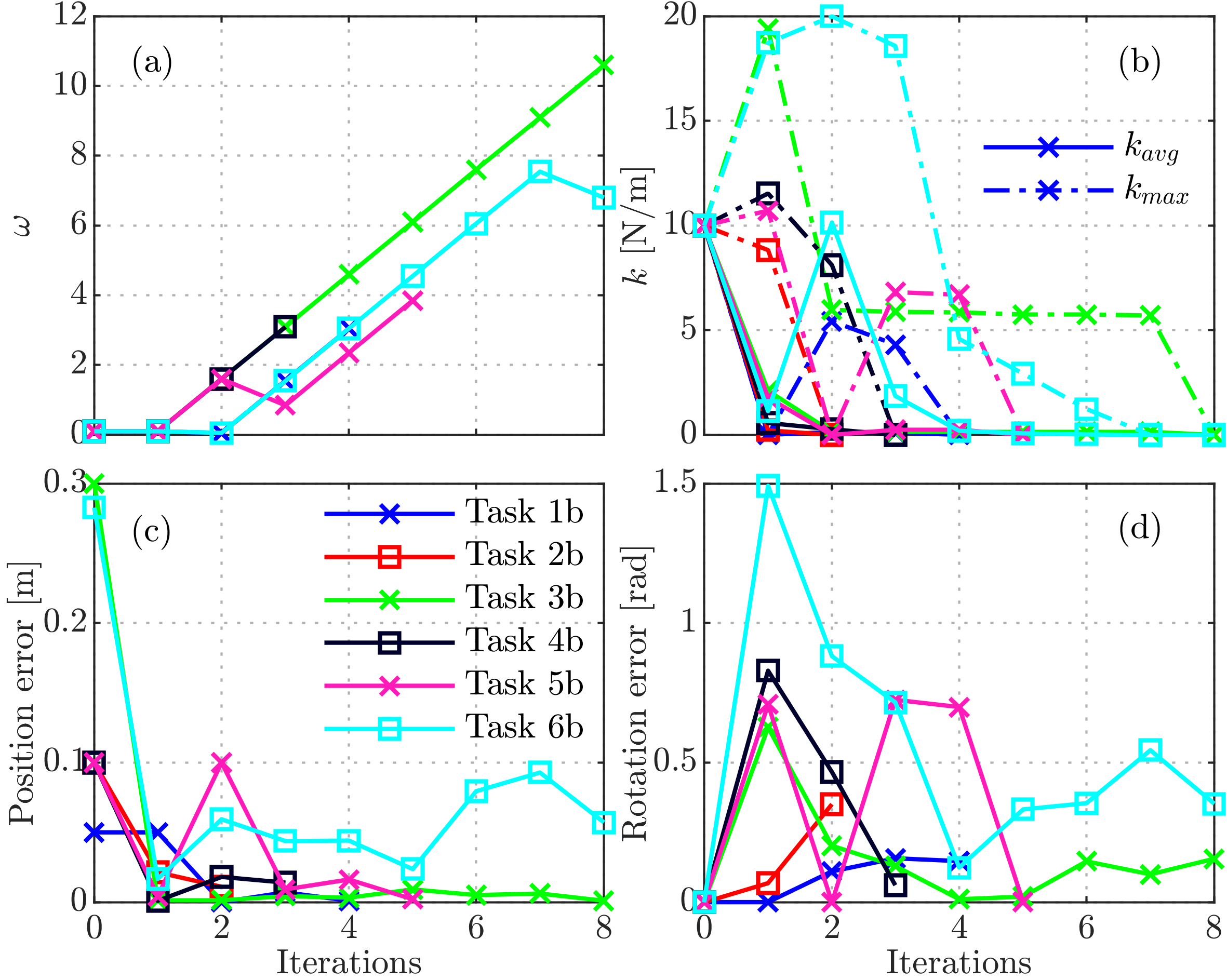}
  \caption{Changes of (a) the penalty, (b) the average and maximum stiffness, (c) position error, and (d) rotation error over penalty iterations for the 7-DOF robot arm tasks.}
  \label{fig:results_sawyer}
  \vspace{-0.12in}
\end{figure}

\begin{figure}
  \centering
  \includegraphics[width=.85\columnwidth]{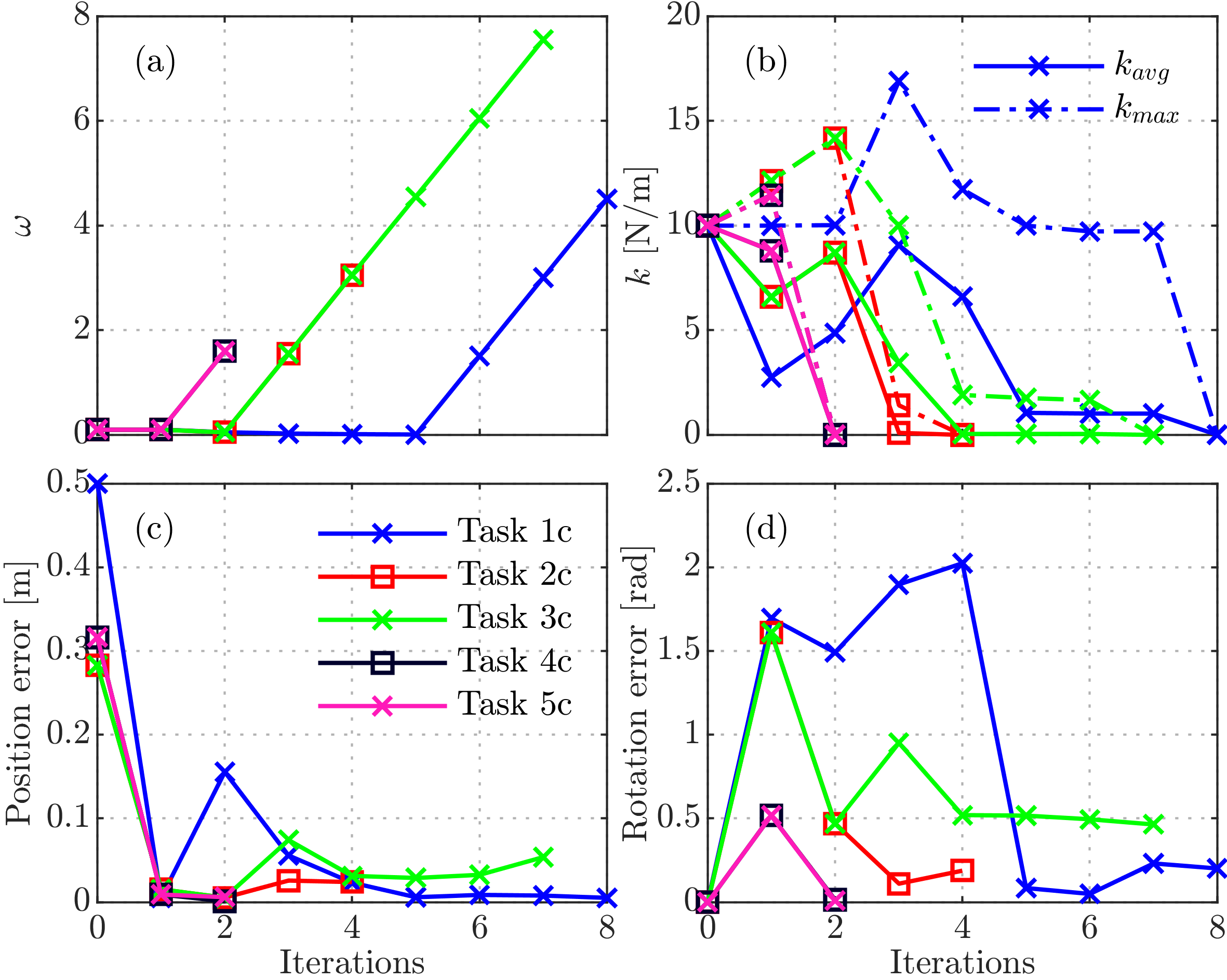}
  \caption{Changes of (a) the penalty, (b) the average and maximum stiffness, (c) position error, and (d) rotation error over penalty iterations for the mobile robot tasks.}
  \label{fig:results_hsr}
\end{figure}

\begin{figure}
  \centering
  \includegraphics[width=.85\columnwidth]{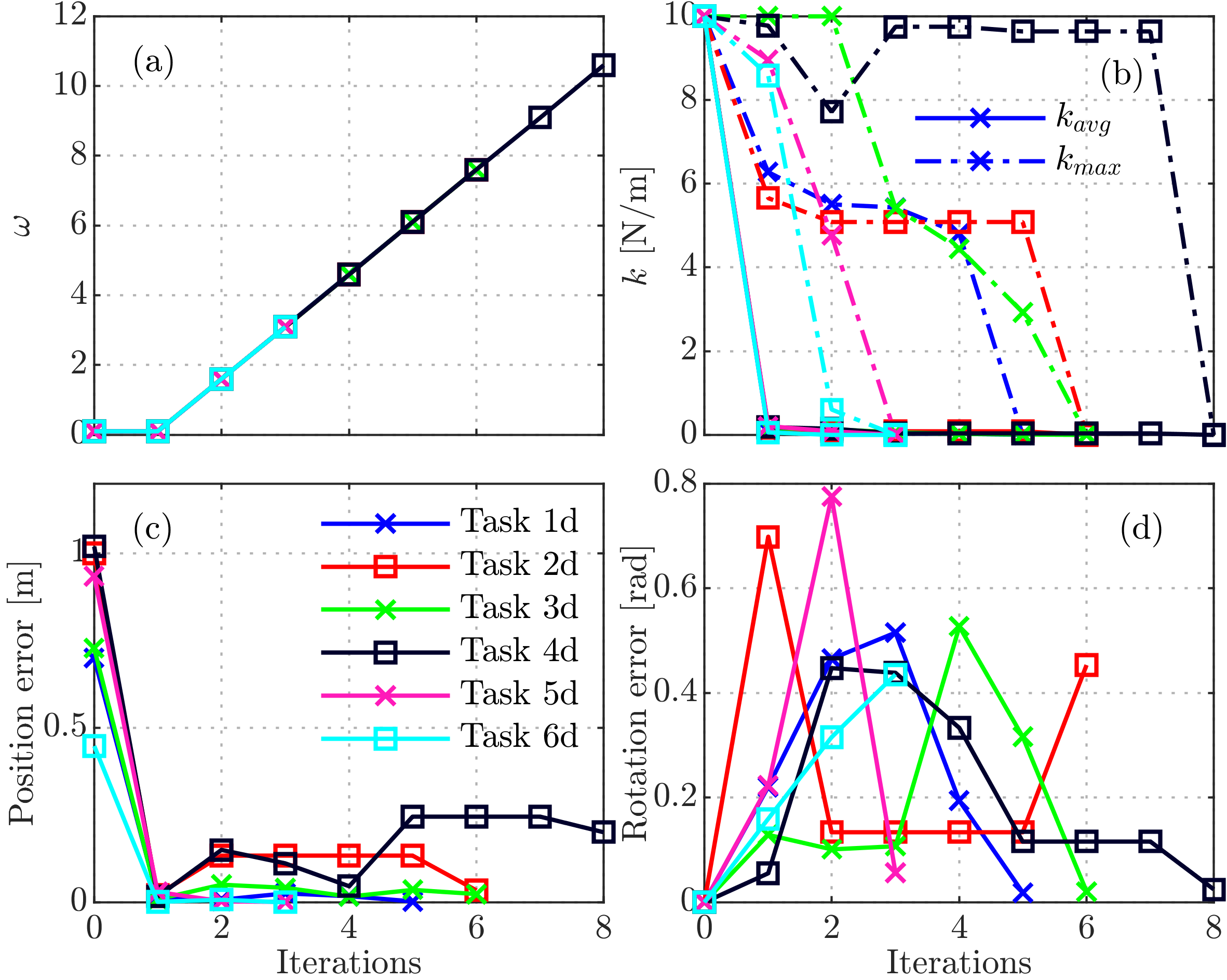}
  \caption{Changes of (a) the penalty, (b) the average and maximum stiffness, (c) position error, and (d) rotation error over penalty iterations for the planar locomotion tasks.}
  \label{fig:results_flymanoid}
  \vspace{-0.12in}
\end{figure}

For all applications, except for the pusher-slider system, we evaluate the method for several goal poses in the workspace. Please see the accompanying video\footnote{The video is available at~\href{https://youtu.be/_GsuxuQEgPg}{https://youtu.be/\_GsuxuQEgPg}.} for the resulting motions. Here, the progress of the algorithm over penalty loop iterations is presented in terms of the penalty over the relaxation, the average and maximum stiffness values, and the pose error. Figures~\ref{fig:results_sawyer}, \ref{fig:results_hsr}, and \ref{fig:results_flymanoid} demonstrate the results for all applications and tasks. These results show that in all cases, the planner finds a motion that satisfies the pose constraints (\textit{i.e.} the normalized position error is below 30\% and the rotation error is below 1 rad) and the stiffness values always converge to zero owing to the exact penalty function in \eqref{eq:ci}. Furthermore, the average stiffness is usually reduced below the threshold in only one iteration.

For the 7-DOF arm tasks, the computation times averaged over penalty loop iterations and tasks for solving the convex subproblems, applying the post process, and calculating the derivatives are 2.05 s, 0.05 s, and 7.23 s, respectively, and the average number of penalty iterations for these tasks is 5. Namely, one can plan a wide range of pushing motions for a 7-DOF arm by using the proposed framework in about 10 s, when the analytic derivatives are integrated.

The video shows that the proposed method can find motions with maintained contacts for sensitive pushing tasks (\textit{e.g.} Task 1b) as well as highly-dynamic behaviors with impact-like contacts (\textit{e.g.} Task 3b), and there are no abrupt changes in the behavior in between. Furthermore, the frictional forces are extensively used for the side pushing tasks although they are not considered in the contact model. This is owing to the fact that once a contact is made, the gradients make it possible to also reason about the tangential forces. Thus, the proposed approach is not limited to the utilization of normal contact forces.

For the mobile robot application, solving the convex subproblems, applying the post process, and calculating the derivatives take 0.91 s, 0.14 s, and 25.81 s in average, and the average number of penalty iterations is 4.6. In this application, the frictional forces are used substantially for the diagonal pushes for the default friction value (\textit{i.e.} Tasks 2c and 4c), as shown in the video. However, lowering the friction coefficient yields similar convergence characteristics, see Fig.~\ref{fig:results_hsr}(a), albeit different motion patterns that mostly rely on normal forces to move the box will occur.

It is noteworthy that the algorithm cannot always find a solution that satisfies the pose constraints in the first iteration (\textit{i.e.} for Tasks 1b, 6b, 1c, 2c, and 3c). However, it overcomes this by reducing the penalty and allowing the virtual stiffness to increase. Additionally, in some cases (\textit{e.g.} Tasks 5b and 6b), the penalty is reduced if the fixed step size $\Delta \omega_s=1.5$ is too large.

The most computationally expensive application is the planar locomotion with the average times of 38.97 s, 0.57 s, and 51.66 s for solving the convex subproblems, applying the post process, and calculating the derivatives and the average number of iterations of 5.2. The similar computation times for the post process indicates that this step scales well with the size of the problem. However, solving the convex program takes a significantly longer time for the locomotion problem, which can potentially be improved by using a customized convex programming solver. Obviously, the generality of the proposed framework comes at the cost of increased computational cost, and real-time implementation remains  unlikely. Nevertheless, it is possible to use the penalty loop approach for offline planning and then executing the resulting trajectory by running SCVX or iLQR in a receding horizon fashion.

\section{Conclusion}
In this paper, we have presented a generalized contact-implicit trajectory optimization framework that can be used for a wide range of applications and tasks with the same trivial initial seed and without any tuning. The proposed algorithm automatically adjusts the penalty on the relaxation parameters while explicitly improving solutions through a post-processing stage that exploits the contact information indicated by how the relaxation is used in the solution. Our findings indicate that this formulation is generic and applicable for a diverse set of problems as it can generate motions with sensitive contact interactions as well as highly-dynamic motions for a wide variety of tasks and robots. The future work will focus on validating the resulting motions by hardware experiments.




\balance
\bibliographystyle{IEEEtran}
\bibliography{refs}

\end{document}